\documentclass[11pt]{article}
\usepackage{ranlp2013}
\usepackage{times}
\usepackage{latexsym}
\usepackage{amsmath}
\usepackage{multirow}
\usepackage{url}
\usepackage{graphicx}
\usepackage{cyrillic}

\title{Analyzing the Use of Character-Level Translation \\with Sparse and Noisy Datasets}

\author{
  J\"org Tiedemann \\
  Department of Linguistics and Philology \\
  Uppsala University \\
  Uppsala, Sweden \\
  {\tt jorg.tiedemann@lingfil.uu.se}\\\And
  Preslav Nakov \\
  Qatar Computing Research Institute\\
  Qatar Foundation, P.O. box 5825\\
  Doha, Qatar\\
  {\tt pnakov@qf.org.qa}
\\}

\date{}

\begin{document}
\maketitle
\begin{abstract}
  This paper provides an analysis of character-level machine
  translation models used in pivot-based translation when applied to
  sparse and noisy datasets, such as crowdsourced movie subtitles. In
  our experiments, we find that such character-level models cut the
  number of untranslated words by over 40\% and are especially
  competitive (improvements of 2-3 BLEU points) in the case of limited training
  data. We explore the impact of character alignment, phrase table
  filtering, bitext size and the choice of pivot language on
  translation quality.  We further compare cascaded translation models
  to the use of synthetic training data via multiple pivots,
  and we find that the latter works significantly better.
  Finally, we demonstrate that neither
  word- nor character-BLEU correlate perfectly with human judgments,
  due to BLEU's sensitivity to length.

\end{abstract}

\section{Introduction}
\label{sec:intro}

Statistical machine translation (SMT) systems, which dominate the
field of machine translation today, are easy to build and offer
competitive performance in terms of translation quality.
Unfortunately, training such systems requires large parallel corpora
of sentences and their translations, called \emph{bitexts}, which are
not available for most language pairs and textual domains.  As a
result, building an SMT system to translate directly between two
languages is often not possible.  A common solution to this problem is
to use an intermediate, or \emph{pivot} language to bridge the gap in
training such a system.

A typical approach is a \emph{cascaded translation} model
using two independent steps of translating from
the source to the pivot and then from the pivot to the target language. A
special
%, and in fact very common,
case is where the pivot
is closely related to the source language, which makes it possible to
train useful systems on much smaller bitexts using
\emph{character-level translation} models. This is the case we will
consider below,
% using the example of
translating Macedonian to English via related languages, primarily Bulgarian.

% , e.g., translating from Macedonian to English
% can be performed using \emph{cascaded translation}:
% first translate from Macedonian to Bulgarian, then from Bulgarian to English.

% %The pivot language is typically chosen so that large bitexts exist for it
% %both to the source and to the target languages;
% %using multiple pivots is also possible.
% %Naturally, the most common pivot language is English since it is rich in
% %both of which are rich in suitable large bitexts.

% Here we revisit the idea by observing that when the pivot language is
% closely related to the source,
% the first translation step requires less work and thus can be carried
% out with much smaller bitexts using \emph{character-level translation} models.

Our main contribution is the further analysis of such a setup. We show
that character-level models can cut the number of untranslated words
almost by half since translation involves many transformations at the
sub-word level. We further explore the impact of character alignment,
phrase table pruning, data size, and choice of pivot
language on the effectiveness of character-level SMT models.
We also study the use of character-level translation for the
generation of synthetic training data, which significantly outperforms
all cascaded translation setups. Finally, we present a manual
evaluation showing that neither word- nor character-BLEU correlate
perfectly with human judgments.

The remainder of the paper is organized as follows:
Section~\ref{sec:related} presents related work.
Section~\ref{sec:char-bases-models} discusses technical details about using character-level SMT models.
% and various extensions thereof that we propose.
%Section~\ref{sec:transliteration} present our character-level transliteration model.
Section~\ref{sec:experiments} describes the experiments,
and Section~\ref{sec:discuss} discusses the results.
Section~\ref{sec:future} concludes with directions for future work.

\section{Related Work}
\label{sec:related}

%TODO: \textbf{We need to find a way to differentiate ourselves from EACL12 and ACL12, which we need to cite here.}

SMT using pivot languages has been studied for several years.
%A number of strategies have been proposed and empirical results have been published for various tasks.
%For example,
\newcite{cohn-lapata:2007:ACLMain} used {\em triangulation}
techniques for the combination of phrase tables. The lexical weights
in such an approach can be estimated by bridging word alignments
\cite{WuWang:2007,Bertoldi:2008:pivot}.

Cascaded translation
%approaches
via pivot languages is used by various researchers
\cite{Gispert:2006,KoehnEA:2009,WuWang:2009}. Several techniques are
compared in \cite{utiyama:isahara:2007:main,Gispert:2006,WuWang:2009}.
%Another strategy is to generate training data using a pivot language
Pivot languages can also be used for paraphrasing and lexical
adaptation \cite{Bannard2005,crego-max-yvon:2010:PAPERS}.  None of
this work exploits the similarity between the pivot and the
source/target language.

% Translation between related languages
The first step in our pivoting experiments involves SMT between closely related languages,
which
%This is arguably simpler than general SMT, and
has been handled using word-for-word translation and manual
%language-specific
rules
%that take care of the necessary morphological and syntactic transformations.
%This has been tried
for a number of language pairs, e.g.,
Czech--Slovak \cite{hajic2000machine},
Turkish--Crimean Tatar \cite{altintas2002machine},
Irish--Scottish Gaelic \cite{scannell2006machine},
Cantonese--Mandarin \cite{Zhang:1998}.
% among others.
%More recently, the Apertium
%open-source machine translation
%platform\footnote{\texttt{http://www.apertium.org/}} has been
%developed, which uses bilingual dictionaries and manual rules to
%translate between a number of related languages, including
%Spanish--Catalan, Spanish--Galician, and Occitan--Catalan, etc.
In contrast, we explore
%purely
statistical approaches that are potentially applicable to many language pairs.

% TODO: Maybe put this back?
% EMNLP'10 -- Thang?
Since we combine word- and character-level models,
a relevant line of research is on combining SMT models of
different granularity, e.g., \newcite{luong-nakov-kan:2010:EMNLP}
combine word- and morpheme-level representations for
English--Finnish.  However, they did not assume similarity
between the two languages, neither did they use pivoting.

Another relevant research combines bitexts between related languages
with little or no adaptation \cite{nakov-ng:2009:EMNLP,Marujo:al:2011,Wang:al:2012,Nakov:Ng:2012}.
However, that work did not use character-level models.

%The use of pivot languages for resource-poor SMT is discussed in \cite{Nakov:2009:ISM}.
Character-level models were used for transliteration
\cite{Matthews:2007,TiedemannNabende:IJCIR09} and for SMT
between closely related languages
\cite{vilar-peter-ney:2007:WMT,Tiedemann:EAMT09a,nakov-tiedemann:2012:ACL}.
% \cite{vilar-peter-ney:2007:WMT,Tiedemann:EAMT09a,tiedemann:2012:EACL,nakov-tiedemann:2012:ACL}.
%Character- and word-level models have been also combined for translating between closely-related languages.
% However, that work did not use pivoting.
\newcite{tiedemann:2012:EACL} used pivoting with character-level SMT.

\section{Character-level SMT Models}
\label{sec:char-bases-models}

Closely related languages
%such as Macedonian and Bulgarian
largely overlap in vocabulary and exhibit strong syntactic and lexical similarities.
Most words have common roots and express concepts with similar linguistic constructions.
%See, for instance, the examples in Figure~\ref{fig.MK-BG-example} of translations between Macedonian and Bulgarian movie subtitles.
Spelling conventions and morphology can still differ,
but these differences are typically regular and thus can easily be generalized.

These similarities and regularities motivate the use of character-level SMT models,
which can operate at the sub-word level, but also cover mappings
spanning over words and multi-word units.
Character-level SMT models, thus combine the
generality of character-by-character transliteration and lexical
mappings of larger units that could possibly refer to morphemes, words or phrases,
to various combinations thereof.

One drawback of character-level models is their inability to model long-distance
word reorderings. However,
%this should not be a strong limitation as
we do not assume very large syntactic differences between closely related languages.
Another issue is that sentences become longer, which
% in the sense
%of longer sequences that need to be translated. This
causes an overhead in decoding time.
% which should be considered.

% Removed: no other subsection
%\subsection{Training Character-level SMT Models}

In our experiments below, we use phrase-based SMT, treating characters as words,
and using a special character for the original space character.
Due to the reduced vocabulary,
%(considering characters instead of words),
we can easily train models of higher order,
thus capturing larger context and avoiding generating non-word sequences:
we opted for models of order 10, both for the language model and for the maximal phrase length (normally, 5 and 7, respectively).
%10-character language model based on our experience from initial experiments.

%As usual, we require a translation model trained on a bitext
%and a target language model trained on monolingual text for
%standard phrase-based SMT.
%Training character-level language models is straightforward using
%the available tools for $n$-gram language modeling.
%The data is further preprocessed as depicted in Figure~\ref{figure.charSMT}.

%Due to the reduced vocabulary (considering characters instead of words),
%we can easily train a model of higher order. It is also necessary to capture
%larger context and avoid generating non-word sequences.
%We opted for a 10-character language model based on our experience from initial experiments.

One difficulty is that training these models requires the alignment of
characters in bitexts. Specialized character-level alignment algorithms do exist,
e.g., those developed for character-to-phoneme translations
\cite{ecs10616,jiampojamarn2007}. However,
\newcite{tiedemann:2012:EACL} has demonstrated that standard tools for
word alignment are in fact also very effective for character-level
alignment, especially when extended with local context. Using
character $n$-grams instead of single characters improves the
expressive power of lexical translation parameters, which are one of
the most important factors in standard word alignment models. For
example, using character $n$-grams increases the vocabulary size of a
1.3M tokens-long Bulgarian text as follows: 101 single characters,
1,893 character bigrams, and 14,305 character trigrams; compared to
30,927 words. In our experiments, we explore the impact of increasing
$n$-gram sizes on the final translation quality. We can confirm that
bigrams perform best, constituting a good compromise between
generality and contextual specificity.

Hence, we used GIZA++ \cite{Och:Ney:2003:mt}
to generate IBM model 4 alignments \cite{Brown:al:1993:mt}
for character $n$-grams,
which we symmetrized using the {\em grow-diag-final-and} heuristics.
%with standard settings
We then converted the result to character alignments by dropping all characters behind the initial one.
%Then, we converted the bigram alignments to character alignments by dropping the second character in each bigram.
%as explained above.
Finally, we used the Moses toolkit \cite{Moses:2007} to build a character-level phrase table.
%we used a maximum phrase length of 10 (characters).

%Finally, the parameters of the log-linear SMT model need to be tuned. This
%is commonly done by minimum error rate training \cite{Och:2003:MERT} using BLEU \cite{Papineni:Roukos:Ward:Zhu:2002},
%or a similar automatic metric, as the objective function to be optimized.

We tuned the parameters of the log-linear SMT model
by optimizing BLEU \cite{Papineni:Roukos:Ward:Zhu:2002}.
%as the objective function to be optimized.
Computing BLEU scores over character sequences does
not make much sense,
especially for small $n$-gram sizes (usually, $n \leq 4$).
Therefore, we post-processed the character-level $n$-best lists
in each tuning step to calculate word-level BLEU.
Thus, we optimized word-level BLEU, while performing character-level translation.
%Thus, we optimized word-level BLEU, while doing character-level decoding
%This ensures that parameters are optimized towards well-performing SMT output of surface strings.
%TODO: \textbf{As proposed in EACL12? Or maybe we should not overcite our papers at submission time?}

\section{Experiments and Evaluation}
\label{sec:experiments}

% \subsection{Datasets and Baselines}

We used translated movie subtitles from the freely available OPUS corpus \cite{Tiedemann:RANLP5}.
The collection includes small amounts of parallel data for Macedonian-English (MK-EN), which we use as our test case.
There is substantially more data for Bulgarian (BG), our main pivot language.
For the translation between Macedonian and Bulgarian, there is even less data available.
%However, this should be big enough for training a character-level translation model.
See Table~\ref{table.dataset}.

\begin{table}[ht]
\centering
{\small
\begin{tabular}{l|cc}
{dataset} & \# sentences & \# words \\
\hline
MK-EN & 160K & 2.2M \\
MK-BG & 102K & 1.3M \\
BG-EN  & 10M & 152M \\
\hline
MK-mono & 536K &   4M\\
BG-mono & 16M & 136M \\
EN-mono &  43M & 435M \\
\hline
\end{tabular}
}
\caption{Size of the datasets.}
% (for both languages together for the parallel data).}
\label{table.dataset}
\end{table}

\noindent The original data from OPUS is contributed by on-line users with little quality control and is thus quite noisy.
Subtitles in OPUS are checked using automatic language identifiers and aligned using time
information \cite{Tiedemann:RANLP5,Tiedemann:LREC12}. However, we identified many misaligned
files and, therefore, we realigned the corpus using \texttt{hunalign}
\cite{VargaEA:2005}.  We also found several Bulgarian files
misclassified as Macedonian and vice versa,
which we addressed by filtering out any document pair for which the BLEU score exceeded 0.7
since it is likely to have large overlapping parts in the same language.
% Bulgarian or Macedonian.
% Preslav: This is confusing: does it mean that we only use files where BLEU is below 0.15?
%BLEU scores for acceptable Macedonian--Bulgarian subtitle pairs
%are otherwise around 0.05 to 0.1 and there were no pairs in the range
%between 0.15 and 0.7.
We also filtered out sentence pairs where the Macedonian/Bulgarian
side contained Bulgarian/Macedonian-specific letters.

From the remaining data we selected 10K sentence pairs (77K English
words) for development and another 10K (72K English words) for
testing; we used the rest for training.  We used 10K pairs because
subtitle sentences are short, and we wanted to make sure that the
dev/test datasets contain enough words to enable stable tuning with
MERT and reliable final evaluation results.

We further used the Macedonian--English and the Bulgarian--English
movie subtitles datasets from OPUS, which we split into dev/test (10K
sentence pairs for each) and train datasets.  We made sure that the
dev/test datasets for MK-BG, MK-EN and BG-EN do not overlap, and that
all dev/test sentences were removed from the monolingual data used for
language modeling.
%Table~\ref{table.dataset} lists the size of the final datasets we used.

% \subsection{Baselines}

Table~\ref{table:baselines} shows our baseline systems,
trained using standard settings for a phrase-based SMT model:
Kneser-Ney smoothed 5-gram language model
and phrase pairs of maximum length seven.
%SRILM to build Kneser-Ney smoothed 5-gram language model,
%Moses to extract and score phrase pairs with a maximum length of seven,
%and MERT to tune the parameters of the overall models.

\begin{table}[ht]
\centerline
{\small
\begin{tabular}{lrrrr}
Task & BLEU & NIST & TER & METEOR \\
\hline
MK-EN   & 22.33  & 5.47 & 63.57 & 39.19 \\% 40.68 52.92 0.99
MK-BG   & 30.70  & 6.52 & 50.94 & 70.44 \\% 77.37 75.78 1.02
BG-MK   & 28.01  & 6.24 & 51.98 & 69.89 \\% 76.50 76.22 1.02
% mk-sr & 31.57  & 6.99 & 49.56 & 51.33 \\%54.58 60.24 0.97
% mk-hr & 39.20  & 8.05 & 40.03 & 53.28 \\%55.23 65.37 1.01
% mk-sl  & 20.27  & 5.29 & 62.25 & 38.45 \\%42.14 51.40 1.02
BG-EN    & 37.60  & 7.34 & 47.41 & 58.89 \\% 61.73 67.11 1.05
\hline
\end{tabular}
}
\caption{Phrase-based SMT baselines.}
\label{table:baselines}
\end{table}

\subsection{Translating Between Related Languages}

% In our first experiments, we will look at translation between related
% languages using character-level SMT models.

% % \subsection{The Impact of Character Alignment and Phrase Table
% % Filtering}

% \subsubsection*{The Impact of Character Alignment and Phrase Table Filtering}

We first investigate the impact of character alignment on the character-level translation between related languages.
For this, we consider the extension of the context using character $n$-grams
proposed by \newcite{tiedemann:2012:EACL}.

%We first investigate the impact of character alignment on the character-level translation between related languages.
%As mentioned earlier, we have several choices between
%alignment approaches and alignment setups. Here, we compare alignments based on weighted finite state transducers (WFSTs)
%resembling edit distance operations with the alignments produced by
%word alignment approaches adapted to the character level. For this, we
%also consider the extension of the context using character $n$-grams
%proposed by \newcite{nakov-tiedemann:2012:ACL}. Note that this approach can also
%be applied to the WFST-based alignment in order to add contextual
%information to the model that is not covered by context-independent
%edit operations.

Another direction we explore is the possibility of reducing the noise in the phrase table.
Treating even closely related languages by transliteration techniques is only a rough
approximation to the translation task at hand. Furthermore, during training we
observe many example translations that are not literally
translated from one language to another.
%% TODO: EXAMPLES somewhere?
Hence, the character-level
phrase table will be filled with many noisy and unintuitive
translation options. We, therefore, applied phrase table pruning
techniques based on relative entropy \cite{Johnson-et-al:2007} to
remove unreliable pairs.

\begin{table}[ht]
\centering
{\small
\begin{tabular}{cr|rr|cc}
   & & \multicolumn{2}{c|}{PT Size} & \multicolumn{2}{c}{BLEU (\%)}\\
\hline
$n$  & align & std & fltd & std & fltd \\
\hline
1 &2.5 &5.1 & 1.0 & 30.47 & 31.13\\
2 &2.6 &5.2 & 0.9 & 30.87 & 31.32\\
3 &2.9 &6.9 & 0.9 & 30.32 & 31.03 \\
4 &3.0 &10.4 & 1.1 & 29.76 & 30.42 \\
5 &3.1 &12.0 & 1.2 & 29.25 & 30.19 \\
6 &3.2 &10.8 & 1.2 & 28.81 & 29.68 \\
7 &3.4 &8.1 & 1.1 & 28.73 & 29.73 \\
8 &3.5 &6.4 & 1.0 & 28.40 & 29.45 \\
9 &3.6 &5.4 & 0.9 & 27.67 & 29.19 \\
10 &3.6&5.1 & 0.9 & 27.11 & 28.78 \\
% 1 &2,465,701 &5,126,108 & 974,477 & 3,001,666 & 7,114,357 & 1,025,646\\
% 2 &2,614,331 &5,205,202 & 936,634 & 3,147,694 & 6,079,822 & 1,023,408\\
% 3 &2,870,179 &6,886,385 & 948,296 & 1,286,939 & 2,089,514 &  708,078 \\
% 4 &2,952,997 &10,371,036 & 1,056,777 & 717,301 & 1,015,876  & 475,541 \\
% 5 &3,059,187 &12,016,323 & 1,189,032 & -- & -- \\
% 6 &3,238,325 &10,805,003 & 1,193,040 & -- & -- \\
% 7 &3,396,915 &8,065,968 & 1,091,520 & -- & -- \\
% 8 &3,506,755 &6,370,621 & 986,812 & -- & -- \\
% 9 &3,576,249 &5,398,241 & 916,146 & -- & -- \\
% 10 &3,601,504&5,064,609 & 872,997 & -- & -- \\
\hline
\end{tabular}
}
\caption{MK-BG character alignment points, phrase table sizes (in million of entries) and BLEU scores
         before (std) and after phrase filtering (fltd).}
\label{tab:pts}
\end{table}

Table~\ref{tab:pts} shows the phrase table sizes for different settings and alignment approaches.
We can see that, in all cases, the size of the filtered phrase tables is less than 20\% of that of the original ones,
which yields significant boost in decoding performance.
More importantly, we see that filtering also leads to consistently better translation quality in all cases.
This result is somewhat surprising for us: in our experience (for word-level models), filtering has typically harmed BLEU.
Finally, we see that both with and without filtering, the best BLEU
scores are achieved for $n=2$.

The numbers in the table imply that the alignments become noisier for $n$-grams longer than two characters;
look at the increasing number of phrases that can be extracted from the aligned corpus,
many of which do not survive the filtering.

%Table~\ref{tab:pt-size} lists the phrase table sizes for different settings and alignment approaches.
%We can see that, in all cases, the size of the filtered phrase tables is less than 20\% of that of the original ones,
%which yields significant boost in decoding performance.
%More importantly, as Table~\ref{tab:char-align} shows,
%filtering also leads to consistently better translation quality in all cases.
%This result is surprising for us: in our experience (for word-level models), filtering has typically harmed BLEU.

% The table shows that GIZA++ alignments become noisier for $n$-grams longer than two characters:
% look at the increasing number of phrases that can be extracted from the aligned corpus
% with many of them not surviving the filtering phase.

%Furthermore, Table~\ref{tab:pt-size} shows a striking difference in the ability to align character $n$-grams.
%The WFST-based aligner does not succeed to align larger $n$-grams (no alignment is found by the
%software in many cases) whereas GIZA++ is very robust in terms of
%alignment generation, which is not very surprising as the alignment
%models are developed for large vocabularies. However, we can see that
%the GIZA++ alignment also becomes noisier for $n$-grams longer than two characters:
%look at the increasing number of phrases that can be extracted from the aligned corpus
%with many of them not surviving the filtering phase.

\subsection{Bridging via Related Languages}

Our next task is to use character-level models in the
translation from under-resourced languages to other languages using
the related language as a pivot.
Several approaches for pivot-based
translations have been proposed as discussed earlier.

We will look at two alternatives: (1)~cascaded translations with two separate
translation models and (2)~bridging the gap by producing synthetic
training corpora. For the latter, we automatically translate the related language in
an existing training corpus to the under-resourced language.

\subsubsection*{Cascaded Pivot Translation}

We base our translations on the individually trained translation models
for the source (Macedonian) to the pivot language (Bulgarian) and for
the pivot language to the final target language (English, in our
case). As proposed by \newcite{tiedemann:2012:EACL}, we rerank $k$-best
translations to find the best hypothesis for each given test
sentence. For both translation steps, we set $k$ to 10 and we require unique translations in the first step.

\begin{table}[ht]
\centerline
{\small
\begin{tabular}{lrrrr}
 & BLEU & NIST & TER & METEOR \\
\hline
Model           & \multicolumn{4}{c}{individually tuned} \\
\hline
word-level pivot   & 22.48  &5.46 &64.11 &47.77\\ %49.16 58.98 1.00
char-based pivot    & 25.67  &5.91 &60.45 &54.61\\ %57.75 62.43 1.02
word+char+MK-EN & 25.00  &5.86 &61.47 &50.19\\ %55.63 57.09 1.01
\hline
Model           & \multicolumn{4}{c}{globally tuned} \\
\hline
word-level pivot   & 23.38  &5.44 &64.33 &48.31\\ %48.83 60.86 1.04
char-based pivot    & 25.73  &5.81 &61.91 &52.47\\ %54.07 62.28 1.03
%w+c pivot    & 26.04 & 5.88 & 61.25 & 52.88\\
word+char+MK-EN    & 26.36  &5.92 &60.85 &53.39\\ %55.70 62.22 1.02
\hline
\end{tabular}
}
\caption{Evaluating cascaded translation: Macedonian to English, pivoting via Bulgarian.}
%  Character-level pivot models use a
%  character-level translation step for translating Macedonian to
%  Bulgarian. The combined model uses three translation paths:
%  word-level pivot, character-level pivot and the direct translation
%  between Macedonian and English.}
\label{table:cascaded}
\end{table}

One possibility is to just apply the models tuned for
the individual translation tasks, which is suboptimal.
% for the global pivot translation.
Therefore, we also introduce a global tuning approach, in which we generate $k$-best lists for the combined
cascaded translation model and we tune corresponding end-to-end weights
using MERT \cite{Och:2003:MERT} or PRO \cite{Hopkins:2011}. % TODO:  (REFERENCES)
We chose to set the size of the $k$-best lists to 20 in both steps to keep
the size manageable, with 400 hypotheses for each tuning sentence.

%We also have a choice about the model for the first step, i.e., the translation from Macedonian to Bulgarian.
%Our main interest is to evaluate the use of character-level SMT models.
%However, we can easily use a word-level model as well, which will serve as a pivot-based baseline in our case.

Another option is to combine  (i) the direct translation model,
(ii) the word-level pivot model, and (iii) the character-level pivot model.
%all with character-level SMT in the first step.
Throwing them all in one $k$-best reranking system does not work well when using the unnormalized model scores.
However, global tuning helps reassign weights such that the interactions between the various components can be covered.
% at least to some extent.
We use the same global tuning model introduced above using a combined
system as the blackbox producing $k$-best lists and tuning feature
weights for all components involved in the entire setup. Using the
three translation paths,
%(direct translation, word-level pivot and character-level pivot)
we obtain an extended set of parameters covering five individual systems.
Since MERT is unstable with so many parameters, we use PRO.
Note that tuning gets slow due to the extensive decoding that is
necessary (five translation steps) and the increased
size of the $k$-best lists (400 hypotheses for each pivot model and 100
hypotheses for the direct translation model).

Table \ref{table:cascaded} summarizes the results of the cascaded
translation models. They all beat the baseline: direct translation from Macedonian to English.
Note that the character-level model adds significantly to the performance of the cascaded model compared
to the entirely word-level one.
%This result illustrates the robustness of the character-level approach.
Furthermore, the scores illustrate that proper weights are important,
especially for the case of the combined translation model. Without globally
tuning its parameters, the performance is below the best single system,
which is not entirely surprising.
%After the global tuning, however, it performs best among all systems.

% Cascaded translation via pivot language:

% \begin{figure*}
% {\footnotesize
% \begin{verbatim}
% pivot:
% - unfiltered word-level
% - filtered char-based, GIZA++ align, 2-gram

% pivot steps:

% (1)    mk-en.word       22.33  5.47 63.57 39.19 40.68 52.92 0.99
% (2a)   mk-bg.word       30.70  6.52 50.94 70.44 77.37 75.78 1.02
% (3a)   mk-bg.char       31.32  6.62 50.11 71.21 77.51 77.09 1.03
% (2/3b) bg-en            37.60  7.34 47.41 58.89 61.73 67.11 1.05

% mk-bg-en.local-word     22.48  5.46 64.11 47.77 49.16 58.98 1.00
% mk-bg-en.local-char     25.67  5.91 60.45 54.61 57.75 62.43 1.02 !!!
% mk-bg-en.local-combined 25.00  5.86 61.47 50.19 55.63 57.09 1.01 !!

% global PRO+MERT:

% mk-bg-en.global-word    23.38  5.44 64.33 48.31 48.83 60.86 1.04 !!
% mk-bg-en.global-char    25.73  5.81 61.91 52.47 54.07 62.28 1.03 !!!
% mk-bg-en.global-combined 26.36  5.92 60.85 53.39 55.70 62.22 1.02 !!!!

% global MERT:

% mk-bg-en.global-word    23.75  5.51 63.46 48.54 49.22 60.71 1.03 !!
% mk-bg-en.global-char    25.64  5.82 61.63 52.39 53.93 62.49 1.04 !!!
% mk-bg-en.local-combined 25.48  5.78 62.28 52.19 53.64 62.62 1.05
% \end{verbatim}
% }
% \end{figure*}

\begin{table}[tbh]
\centerline
{\small
\begin{tabular}{l@{ }rrr@{ }r}
Model & BLEU & NIST & TER & METEOR \\
\hline
BG word-syn.            & 26.01  & 5.82 & 61.49 & 50.31 \\%50.55 62.41 1.03 !
BG char-syn.            & 28.17  & 6.17 & 58.97 & 55.27 \\%57.10 64.11 1.02 !!
BG w+c-syn.             & 28.62  & 6.25 & 58.52 & 55.75 \\%57.88 64.07 1.02 !!!
BG w+c-syn.+MK-EN       & 29.11  & 6.30 & 58.27 & 56.64 \\
\hline
\hline
SR word-syn.            & 25.39  & 5.72 & 63.26 & 41.15 \\
SR char-syn.            & 27.25  & 6.14 & 60.31 & 47.32 \\
SR w+c-syn.             & 29.05  & 6.29 & 59.73 & 49.18 \\
SR w+c-syn.+MK-EN       & 30.39  & 6.51 & 58.08 & 50.91 \\
\hline
SL word-syn.            & 24.78  & 5.58 & 64.04 & 39.34 \\
SL char-syn.            & 24.03  & 5.67 & 63.11 & 44.76 \\
SL w+c-syn.             & 27.30  & 6.11 & 60.46 & 47.69 \\
SL w+c-syn.+MK-EN       & 28.42  & 6.26 & 59.57 & 49.06 \\
\hline
CZ word-syn.            & 26.48  & 5.83 & 62.52 & 41.02 \\
CZ char-syn.            & 23.74  & 5.51 & 64.96 & 44.96 \\
CZ w+c-syn.             & 28.03  & 6.08 & 61.12 & 48.41 \\
CZ w+c-syn.+MK-EN       & 29.24  & 6.29 & 59.39 & 49.60 \\
\hline
ALL-syn.                & 36.25 & 7.24 & 53.06 & 61.74 \\
ALL-syn.+MK-EN          & 36.69  & 7.28 & 52.83 & 62.26 \\
\hline
\end{tabular}
}
\caption{Macedonian-English translation using synthetic data (by translating $X$-EN to MK-EN).}
%using different pivot languages: Bulgarian (BG), Serbian (SR), Slovenian (SL), and Czech (CZ).}
%  The training data is created by translating Bulgarian to Macedonian in a
%  large Bulgarian-English parallel corpus
%  using a {\em word-level} model, a {\em char-based} model. {\em
%    Combined} uses a concatenation of both.}
\label{table:synthetic}
\end{table}

\subsubsection*{Synthetic Training Data}

Another possibility to make use of pivot languages is to create synthetic training data.
For example, we can translate the Bulgarian side of our large Bulgarian--English training bitext to Macedonian,
thus ending up with ``Macedonian''-English training data.
This is similar to previous work on adapting between closely related languages \cite{Marujo:al:2011,Wang:al:2012},
but here we perform translation rather than adaptation.

%Another possibility is to create synthetic training data.
%% for under-resourced language pairs.
%In our case, we will use models for
%closely related languages (Macedonian and Bulgarian) to translate existing training data for a
%better supported language pair (here Bulgarian--English).
%
%In order to prepare a system for the translation from Macedonian to
%English, we translated the Bulgarian part of our large
%Bulgarian--English training corpus to Macedonian.

For translating from Bulgarian to Macedonian,
%(1)~translate using a word-level model and
%(2)~translate using a character-level model.
we experimented with a word-level and a character-level SMT model.
We also combined the two by concatenating the resulting MK-EN bitexts.
We relied on the single best translation in all cases.
Here, the character-level model was our best performing one,
when using a filtered phrase table based on bigram alignments.
Using $k$-best lists would be another option,
but those should be properly weighted when combined to form a new synthetic training set.
In future work, we plan to try the more sophisticated bitext combinations
from \cite{nakov-ng:2009:EMNLP}.

%% TODO: add performance of the bg-mk char-model here

Table~\ref{table:synthetic} shows the overall results.
Experiments with word-level and character-level SMT are shown in rows
1 and 2, respectively.
% We experimented with a word-level and a character-level SMT model (e.g., rows 1-2).
The result from the model trained on a concatenation of synthetic
bitexts is shown in the third row.
% We also combined the two by concatenating the resulting MK-EN bitexts from each of them (e.g., row 3).
Finally, we also added the original MK-EN bitext to the combination (e.g., row 4).
We can see from the top four rows that synthetic data outperforms cascaded translation by 2-3 BLEU points.
%for using Bulgarian as a pivot are presented in the first four lines of Table~\ref{table:synthetic}.
%As we can see, using synthetic data works surprisingly well.
%Translating our related languages seems to be robust enough to create a reasonable training
%corpus to train an SMT system for our target language pair
%that outperforms all our other systems significantly.
Here again, the character-level model is much more valuable than the word-level one,
which is most probably due to the reduction in the number of out-of-vocabulary (OOV)
words it yields.

Another huge advantage over the cascaded
approaches presented above is the reduced decoding time. Now, the
system behaves like a traditional phrase-based SMT engine. The only
large-scale effort is the translation of the training corpus,
which only needs to be done once and can easily be performed off-line in a distributed setup.

\subsection{Learning Curves and Other Languages}

Observing the success of bridging via related pivot languages leads to
at least two additional questions:
(1)~How much data is necessary for
training reasonable character-level translation models
that are still better than a standard word-level model trained on the same data?
and
(2)~How strongly related should the languages be
so that it is beneficial to use SMT at the character level?

\subsubsection*{Size of the Training Data}

We investigated the first question
by translating from Macedonian to Bulgarian with increasing amounts of training data.
For comparability, we kept the model parameters fixed.
%(feature weights)
%of our systems the same.
%(word-level baseline, best character-level model).

\begin{figure}[htbp]
 \hspace*{-0.8cm}\mbox{
   \includegraphics[width=0.3\textwidth]{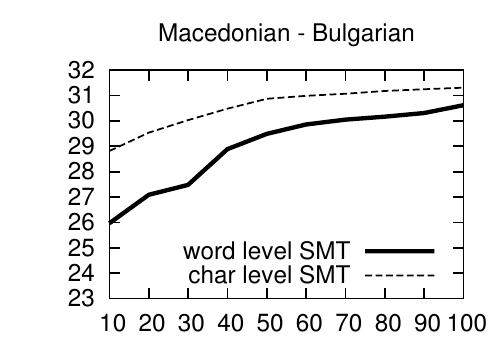}
   \hspace{-0.9cm}
   \includegraphics[width=0.3\textwidth]{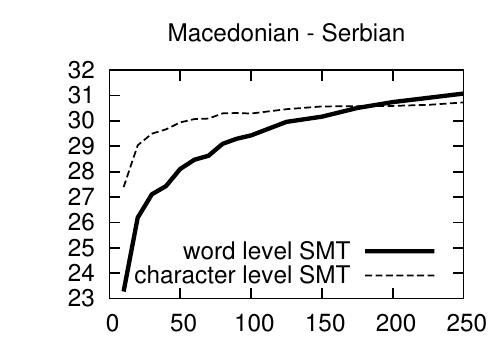}
 }
 \hspace*{-0.8cm}\mbox{
   \includegraphics[width=0.3\textwidth]{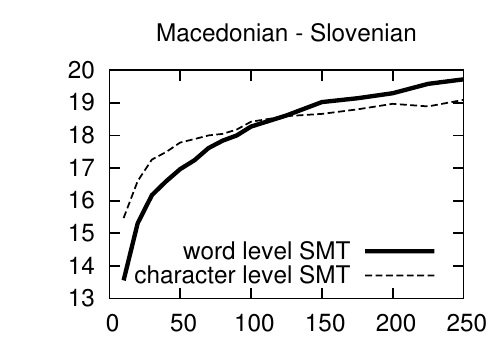}
   \hspace{-0.9cm}
   \includegraphics[width=0.3\textwidth]{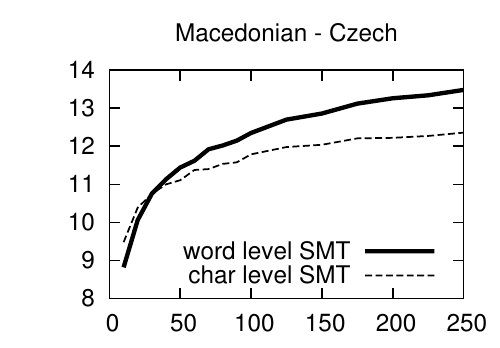}
 }
 \caption{BLEU (in \%) for word- and character-level SMT models
   with varying sizes of parallel training data (in thousands of
   sentence pairs).}
 \label{img:learning_mkbgsrsl} %
 \end{figure}

The top-left plot in Figure~\ref{img:learning_mkbgsrsl}
%Figure~\ref{img:learning_mkbg}
shows the learning curves for word- and character-level models for MK-BG.
We can see that the character-level models clearly outperform the
word-level ones for the small amounts of training data that we have:
the abstraction at the character level is much stronger and yields more robust models.
%However, the word-level model does not reach the performance
%of the character models at any time in our setup which is mainly due
%to the lack of more training data in our case.

% \begin{figure}[htbp]
% \centering
% %\includegraphics[width=0.45\textwidth]{mk-bg}
% \includegraphics[width=0.30\textwidth]{mk-bg-100}
% \caption{BLEU (in \%) for word-level and character-level SMT models with varying sizes of
%    parallel training data (in thousands of sentence pairs).}
% % \caption{Learning curves for word- and character-level MK-BG models for varying sizes of data.}
% \label{img:learning_mkbg} %
% \end{figure} %

\subsubsection*{Other Pivot Languages}

%For further investigations, we need larger training sets which we do
%not have available from the subtitle domain for
%Macedonian-Bulgarian. Furthermore, we would also like to look at the
%second question about relatedness of the languages.

We investigated the second question
by experimenting with data from OPUS for two South-Slavic languages that are less related to Macedonian than Bulgarian.
We selected Serbian and Slovenian from the Western group of the South-Slavic language branch
(Bulgarian and Macedonian are in the Eastern group) from which Slovenian is the furthest away from Macedonian.

Note that while Bulgarian and Macedonian use Cyrillic,
Slovenian and Serbian use the Latin alphabet (Serbian can also use Cyrillic, but not in OPUS).
We have larger training datasets for the latter two: about 250-300 thousand sentence pairs.
%We trained character-level and word-level SMT models in the
%same way we did for Macedonian-Bulgarian, again using different
%training sizes.

To further contrast the relationship between South-Slavic languages (such as Bulgarian, Macedonian, Serbian, Slovenian)
and languages from the Western-Slavic branch,
we also experimented with Czech (about 270 thousand sentence pairs).

Note that we do not have the same movies available in
all languages involved; therefore, the test and the development datasets
are different for each language pair. However, we used the same
amount of data in all setups: 10,000 sentence pairs for tuning and 10,000 pairs for evaluation.
%Of course, the absolute scores are not comparable between language pairs.

% \begin{figure}[htbp]
% \hspace{-0.5cm}\mbox{
% \includegraphics[width=0.27\textwidth]{mk-sr}
% \hspace{-0.5cm}
% \includegraphics[width=0.27\textwidth]{mk-sl}
% }
% \caption{Training word-level and character-level SMT models for
%   Macedonian-Serbian and Macedonian-Slovenian with varying sizes of
%   parallel training data.}
% \label{img:learning_mksrsl} %
% \end{figure} %

%  \begin{figure*}[htbp]
% \begin{center}
%  \hspace{-0.5cm}\mbox{
%  \includegraphics[width=0.3\textwidth]{mk-sr}
%  \hspace{-0.5cm}
%  \includegraphics[width=0.3\textwidth]{mk-sl}
%  \hspace{-0.5cm}
%  \includegraphics[width=0.3\textwidth]{mk-cs}
%  }
% \end{center}
%  \caption{BLEU (in \%) for word-level and character-level SMT models for
%    Macedonian-Serbian and Macedonian-Slovenian with varying sizes of
%    parallel training data (in thousands of sentence pairs).}
%  \label{img:learning_mksrsl} %
%  \end{figure*} %

Figure~\ref{img:learning_mkbgsrsl} also shows the learning curves for
the three additional language pairs.  For the South-Slavic languages,
the character-level models make sense with sparse datasets.
They outperform word-level models at least until around 100 thousand
sentence pairs of training data.
%in our domain.

Certainly, language relatedness has an impact on the effectiveness of
character-level SMT. The difference between character- and
word-level models shows that Slovenian is not the most appropriate
choice for character-level SMT due to its weaker relation to Macedonian.

Furthermore, we can see that the performance of
character-level models levels out at some point and standard word-level models
surpass them with an almost linear increase in MT quality up to the
point we have considered in the training procedures. Looking at Czech, we
can see that character-level models are only competitive for very small
datasets, but their performance is so low that they are practically useless.
In general, translation is more difficult for more distant languages;
this is also the case for word-level models.

\subsubsection*{Pivot-Based Translation}

The final, and probably most important, question with respect to this
paper is whether the other languages are still useful for pivot-based
translation. Therefore, we generated translations of our
Macedonian-English test set but this time via Serbian, Slovene and
Czech. We used the approach that uses synthetic training data, which was
the most successful one for Bulgarian, based on translation models
trained on subsets of 100 thousand sentence pairs to make the results
comparable with the Bulgarian case.

% For simplicity, we used the cascaded two-step approach without
% step-specific parameter tuning. Furthermore, we used models trained
% on a subset of 100 thousand sentence pairs for the translation from
% Macedonian to the pivot language to make the results comparable with
% the Bulgarian case.

The pivot--English data that we have translated to ``Macedonian''-English
%available for the training systems from the pivot language to English
is comparable for Slovene and Serbian (1 million sentence
pairs) and almost double the size for Czech (1.9 million).

Table~\ref{table:synthetic} summarizes the results for all pivot languages:
Bulgarian (BG), Serbian (SR), Slovenian (SL), and Czech (CZ).
%Serbian (SR) and Slovenian (SL), which are from the Western branch of the South-Slavic subgroup of the Slavic language group
%(Bulgarian and Macedonian are in the Eastern branch),
%and Czech (CZ), which is from the Western Slavic subgroup;
% We had 100K training sentence pairs for all Macedonian-pivot models.
We can see that Serbian, which is geographically adjacent to Macedonian,
performs almost as well as Bulgarian, which is also adjacent,
while Slovenian, which is further away, and not adjacent to any of the above, performs worse.
Note that with a Slovenian pivot, the character-level model performs worse than the word-level model.

This suggests that the differences between Slovenian and Macedonian are not that much at the sub-word level but mostly at the word level.
This is even more evident for Czech, which is geographically further away and which is also from a different Slavic branch.
Note, however, that we had much more data for Czech-English than for any other $X$-EN bitext,
which explains the strong overall performance of its word-level model.

Overall, we have seen that as the relatedness between the source and the pivot language decreases,
so does the utility of the character-level model.
However, in all cases the character-level model helps when combined with the word-level one,
%\footnote{Note that
%here we only combine word- with character-level models; we do NOT use the MK-EN bitext.}
yielding 1.5-4 BLEU points of improvement.
%i.e., even with less related languages, character-level models are useful.
Moreover, using all four pivots yields seven additional BLEU points over the best single pivot.

\section{Discussion}
\label{sec:discuss}

Finally, we performed a manual evaluation of word-level, character-level and combined systems
translating from Macedonian to English using Bulgarian.
We asked three speakers of Macedonian and Bulgarian
%especially at the level of spoken language, which is used in movie subtitles.}
to rank the English output from the eight anonymized systems in Table~\ref{table:manual:eval},
given the Macedonian input; we used 100 test sentences.

\begin{table}[ht]
\centerline
{\small
\begin{tabular}{@{ }@{ }lccccc@{ }@{ }}
      & word & char & Avg. & ``$>$'' & Untr.\\
Model & BLEU & BLEU & rank & score & words\\
\hline
reference  & --- & --- & 1.57  &  0.73 & ---\\
\hline
baseline   &  22.33 & 50.83 & 3.37  &  0.25 & 4,959\\
\hline
word-pivot  &  23.38 & 53.26 & 2.81  &  0.42 & 3,144\\
char-pivot  &  25.73 & 56.00 & 2.51  &  0.52 & 1,841\\
comb-pivot  &  26.36 & 56.39 & 2.63  &  0.46 & 1,491\\
\hline
word-synth.  &  26.01 & 55.59 & 2.77  &  0.43 & 3,258\\
char-synth.  &  28.17 & 58.21 & 2.31  &  0.58 & 1,818\\
comb-synth.  &  28.62 & 58.53 & 2.11  &  0.65 & 1,712\\
\hline
\end{tabular}
}
\caption{Comparing word- and character-level BLEU
         to human judgments for MK-EN using BG.}
%         Macedonian-English using Bulgarian:
%         global pivoting vs. synthetic data (see Tables \ref{table:cascaded} and \ref{table:synthetic}).}
\label{table:manual:eval}
\end{table}

The results are shown in Table~\ref{table:manual:eval}.
Column 4 shows the \emph{average rank} for each system,
and column 5 shows the \emph{``$>$'' score} as defined in \cite{callisonburch-EtAl:2012:WMT}:
the frequency a given system was judged to be strictly better than the rest
divided by the frequency it was judged strictly better or strictly worse than the rest.
We further include word- and character-level BLEU,
and the number of untranslated words.

%Note that the reference was not always ranked first,
%i.e., the SMT systems often managed to get better, e.g., more literal, translations.
%% TODO add reference to issues with translation direction

%The agreement between the individual annotators at the system-pair level was very high.
We calculated Cohen's kappas \cite{Cohen:kappa} of 0.87, 0.86 and 0.83 between the pairs of judges,
following the procedure in \cite{callisonburch-EtAl:2012:WMT}.
This corresponds to almost perfect agreement \cite{landis1977},
probably due to the short length of subtitles, which allows for few differences in translation and simplifies ranking.

The individual human judgments (not shown to save space) correlate perfectly
%between themselves
in terms of relative ranking of (a)~the three pivoting systems and (b)~the three synthetic data systems.
Moreover, the individual and the overall human judgments also correlate well with the BLEU scores on (a) and (b),
with one notable exception: humans ranked \textit{char-pivot} higher than \textit{comb-pivot},
while word- and char-BLEU switched their ranks.
A closer investigation found that this is probably due to length:
the hypothesis/reference ratio for \textit{char-pivot} is 1.006, while for \textit{comb-pivot} it is 1.016.
In contrast, for \textit{char-synth.} it is 1.006, while for \textit{comb-synth.} it is 1.002.
%The closer it is to 1, the better the translations \cite{Nakov:al:2012:COLING}.
Recent work \cite{Nakov:al:2012:COLING} has shown that the closest this ratio gets to 1,
the better the BLEU score is expected to be.
%this would explain the BLEU scores.

Note also that word-BLEU and char-BLEU correlate perfectly on (a) and (b),
which is probably due to tuning the two systems for word-BLEU.
%we could have seen a different picture if the character-level systems had optimized character-BLEU.

Interestingly, the BLEU-based rankings of the systems inside (a) and
(b) perfectly correlate with the number of untranslated words.
Note the robustness of the character-level models: they reduce the number of untranslated words by more than 40\%.
Having untranslated words in the final English translation could be annoying since they are in Cyrillic,
but more importantly, they could contain information that is critical for a human,
or even for the SMT system, without which it could not generate a good translation for the remaining words in the sentence.
This is especially true for content-baring long, low-frequency words.
For example, the inability to translate the Macedonian
%\footnote{While most Cyrillic words in the translation originate in the Macedonian input,
%in one case, we found the Bulgarian word \textcyrrm{kalao} instead of the Macedonian \textcyrrm{kaleo}.}
\textcyrrm{lutam} (`I am angry') yields ``You don' \textcyrrm{lutam}.'' instead of ``I'm not mad at you.''

Character models are very robust with unknown morphological forms,
e.g., a word-level model would not translate \textcyrrm{razveselam}, yielding ``I'm trying to make a \textcyrrm{razveselam}.'',
while a character-level model will transform it to the Bulgarian \textcyrrm{razveselya},
thus allowing the fluent ``I'm trying to cheer you up.''
Note that this transformation does not necessarily have to pick the correct Bulgarian form,
e.g., \textcyrrm{razveselya} is a conjugated verb (1st person, singular, subjunctive),
but it is translated as an infinitive,
i.e., all Bulgarian conjugated forms would map to the same English infinitive.

Finally, it is also worth mentioning that character models are very robust
in case of typos, concatenated or wrongly split words,
which are quite common in movie subtitles.

\section{Conclusion and Future Work}
\label{sec:future}

We have explored the use of character-level SMT models when
applied to sparse and noisy datasets
such as crowdsourced movie subtitles.
%which often contain many typos and wrong word concatenations and segmentations.
We have demonstrated their utility when translating between closely related languages,
where translation is often reduced to sub-word transformations.
We have shown that such models are especially competitive in the case of limited training data
(2-3 BLEU points of improvement, and 40\% reduction of OOV),
but fall behind word-level models as the training data increases.
We have
also shown the importance of phrase table filtering and the impact of
character alignment on translation performance.

We have further experimented with bridging via a related language and
we have found that generating synthetic training data works best.
This makes it also straightforward to use multiple pivots and to combine
word-level with character-level SMT models. Our best combined model outperforms the
baseline by over 14 BLEU points, which represents a very significant boost in
translation quality.
% for our under-resourced language at hand.
%;
%surprisingly, we found that using synthetic data works better (by 2-3 BLEU points).

In future work, we would like to investigate the robustness of
character-level models with respect to domain shifts and for other language pairs.
We further plan a deeper
analysis of the ability of character-level models to handle noisy
inputs that include spelling errors and tokenization mistakes,
which are common in user-generated content.

%we have revisited cascaded translation using a pivot language for the special case when the pivot is closely related to the source language,
%and we have compared it to using synthetic data, which turned out to work better.
%Overall, we have found that character-level models are especially competitive in the case of limited training data,
%but fall behind word-level models as the training data increases.
%We have further explored the impact of the choice of the pivot language and the effect of domain shifts.
%We used both automatic and manual evaluation, and we noticed that BLEU is very sensitive to length.

% In future work, we plan to combine pivoting with synthetic data,
% e.g., using more complex global optimization or a simple system combination.
% The robustness of character-level models to noisy input
% also suggests trying SMT without tokenization,
% which would eliminate the danger of not being able to translate due to wrong tokenization.

% TODO: Thank Veno for the evaluation in the final paper.

\vspace{-4pt}
\section*{Acknowledgments}
\vspace{-2pt}
We would like to thank Petya Kirova and Veno Pacovski
for helping us with the manual judgments.
We also thank the anonymous reviewers for their constructive comments.
%which have helped us improve the paper.
% Any EU projects to cite?

\bibliographystyle{acl}

\bibliography{bibliography}

\end{document}